\documentclass[10pt, conference, compsocconf]{IEEEtran}

\usepackage{cite}
\usepackage{algorithmic}
\usepackage{url}
\usepackage[cmex10]{amsmath}
\usepackage{amsfonts}
\usepackage{booktabs}
\usepackage{array}
\usepackage{multirow}
\usepackage{graphicx}

\newcommand{\argmax}{\operatornamewithlimits{argmax}}
\newcommand{\argmin}{\operatornamewithlimits{argmin}}

\begin{document}
\title{An Empirical Study of MAUC in Multi-class Problems with Uncertain Cost Matrices}
\author{Rui Wang and Ke Tang\\
wrui1108@mail.ustc.edu.cn, ketang@ustc.edu.cn\\
Nature Inspired Computation and Applications Laboratory (NICAL), School of Computer Science\\
and Technology, University of Science and Technology of China, Hefei 230027, China\\
}

\maketitle

\begin{abstract}
Cost-sensitive learning relies on the availability of a known and fixed cost matrix.  However, in some scenarios, the cost matrix is uncertain during training, and re-train a classifier after the cost matrix is specified would not be an option. For binary classification, this issue can be successfully addressed by methods maximizing the Area Under the ROC Curve (AUC) metric. Since the AUC can measure performance of base classifiers independent of cost during training, and a larger AUC is more likely to lead to a smaller total cost in testing using the threshold moving method. As an extension of AUC to multi-class problems, MAUC has attracted lots of attentions and been widely used. Although MAUC also measures performance of base classifiers independent of cost, it is unclear whether a larger MAUC of classifiers is more likely to lead to a smaller total cost. In fact, it is also unclear what kinds of post-processing methods should be used in multi-class problems to convert base classifiers into discrete classifiers such that the total cost is as small as possible. In the paper, we empirically explore the relationship between MAUC and the total cost of classifiers by applying two categories of post-processing methods. Our results suggest that a larger MAUC is also beneficial. Interestingly, simple calibration methods that convert the output matrix into posterior probabilities perform better than existing sophisticated post re-optimization methods.
\end{abstract}


\section{Introduction}
In many classification problems, different types of misclassifications would cause different cost, and the target of classification is to minimize the total cost. This problem has been extensively investigated in the context of cost-sensitive learning \cite{Elkan2001}. However, most existing work presume that the misclassification cost matrix is specified in advance \cite{Domingos_1999, Ting_2002, Zhou2006}, while this is never guaranteed in the real world. More importantly, once a classifier is built, it may be used in an uncertain environment, i.e., the cost matrix may change over time \cite{Provost2001}. For example, in the stock market, the cost of different wrong predictions could change frequently with time. On the other hand, the training data is too huge or expensive for re-train the classifier every time when the cost is changed. In such a situation, one might want to execute the training procedure only once and reuse the classifier for changed cost.

For binary problems, this issue can be well addressed by methods that maximizing the Area Under the ROC Curve (AUC) \cite{Fawcett2006}. That is, one can firstly train a base classifier which output numerical scores for input instances with large AUC. Once the cost matrix is specified, the base classifier is converted into a discrete classifier to assign class labels to input instances. This procedure is well justified since it has been proved that AUC measures the performance of classifiers independent of cost and a classifier with larger AUC would more likely be converted to a classifier with smaller total cost \cite{Fawcett2006}. Moreover, there is a simple and effective method (i.e., the threshold moving method) to implement the conversion. Hence, when the cost matrix changes, one only needs to execute the conversion again instead of re-training the classifier. However, in spite of its nice properties and big success, AUC can only be applied to binary problems. One simple and widely used  \cite{Huang2005,Tang2011,Zhou2010,Cerquides2005,Sikonja2004} generalization of it for multi-class problems is the so called MAUC metric \cite{Hand2001}. Generally speaking, MAUC is the average of the AUC of each pair of classes. Therefore, it can also measure the performance of classifiers independent of cost.

In \cite{Huang2005}, Huang and Ling compared the AUC/MAUC with accuracy. It is found that AUC/MAUC is statistically consistent with accuracy and statistically more discriminant than accuracy. However, in the context of minimizing total cost, it is more important to verify whether there exists negative correlation between the MAUC and the total cost of classifiers. Unfortunately, despite the evidence that a larger AUC can more likely to lead to a smaller total cost in binary problems, there is neither theoretical guarantees nor empirical studies published in the literature to justify the preassumption that a larger MAUC can also more likely to lead to a smaller total cost. Without such a justification, the adoption of MAUC as performance metric is fragile.

To explore the relationship between the MAUC and the total cost of classifiers, one has to firstly address how to convert a base classifier to a discrete classifier such that the total cost is as small as possible. For binary problems, there is only one pair of classes, so the simple threshold moving method can find the minimum total cost effectively by scanning over all possible discrete classifiers \cite{Sheng_Ling_2003}. However, for multi-class problems, the output of the base classifier is a matrix. There is no effective way to find the discrete classifier with minimum total cost \cite{Lachiche2003}. Therefore, several researchers have studied how to re-optimize the output matrix explicitly to obtain small cost for specified cost matrix. Lachiche and Flach \cite{Lachiche2003} formulated this conversion task as an optimization problem, and proposed a hill-climbing heuristic method.  Bourke et al. \cite{Bourke2008} furthermore proved that the decision version of this formulated problem is NP-complete and experimentally compared several heuristic re-optimization methods. In summary, they found that the genetic algorithm (GA), the method proposed by Lachiche and Flach \cite{Lachiche2003} (LF) and the MetaClass method are better than the other compared methods. In addition to these re-optimization methods, an alternative is to transform the output scores of base classifiers into posterior probabilities. If the posterior probabilities are sufficiently accurate, the decisions with the smallest cost can be made according to the Bayesian rule \cite{duda2001pattern}. These type of methods are referred to as output calibration methods, and are usually designed for more general purposes.

This paper provides an empirical study to answer the question whether a large MAUC is more useful in multi-class problems with uncertain cost matrices. Through extensive experiments on 26 data sets and 5 different types of classification algorithms, it is found that, when equipped with good conversion methods, a classifier with larger MAUC can indeed more likely to lead to a smaller total cost. Moreover, although the explicit conversion methods appear to have attracted more attentions recently, they were outperformed by the calibration methods in our experiments.

The rest of the paper is organized as follows. After the MAUC metric is briefly introduced in Section \uppercase\expandafter{\romannumeral2}, three re-optimization methods compared in \cite{Bourke2008} and two simple and widely used calibration methods are reviewed in Section \uppercase\expandafter{\romannumeral3}. The details of our experiments and the analysis of their results are presented in Section \uppercase\expandafter{\romannumeral4}. In Section \uppercase\expandafter{\romannumeral5}, we end the paper with some conclusions and discussions.

\section{MAUC}
Given a data set $S=\{(x_1, y_1), (x_2, y_2), \ldots, (x_n, y_n)\}$, where $x_i\in{\mathbb{R}^m}$ and $y_i\in\{1, 2, \ldots, c\}$ are the feature vector and class label of the \emph{i-th} instance respectively. A classifier $H(x_i)\rightarrow{\mathbb{R}^c}$ assigns a score vector to instance $x_i$, each element of the vector indicates the confidence that instance $x_i$ belongs to the corresponding class. Therefore, the output of classifier $H$ for data set $S$ can be arranged as a $n\times{c}$ matrix. Then the MAUC of the classifier can be calculated \cite{Hand2001}.
\begin{equation}
  \text{\emph{MAUC}}=\frac{2}{c\times(c-1)}\sum_{i<j}{\frac{A_{ij}+A_{ji}}{2}}
  \label{mauc}
\end{equation}
where $A_{ij}$ ($A_{ji}$) is the AUC value calculated from the \emph{i-th} (\emph{j-th}) column of the output matrix considering instances from class \emph{i} and \emph{j}, notice that they are not necessary equal to each other \cite{Hand2001}. Furthermore, the AUC value of a classifier on binary problem can be calculated as,
\begin{equation}
  \text{\emph{AUC}}=\frac{\sum_{x_i\in{class(+)};\,x_j\in{class(-)}}{s(x_i,x_j)}}{n^+\times{n^-}}
\end{equation}
where $n^+$ ($n^-$) is the number of instances in positive (negative) class. Using $h(x)$ to indicate the confidence that $x$ belongs to positive class, then $s(x_i, x_j)$ is defined as,
\begin{equation}
  s(x_i,x_j)=
  \left\{
    \begin{array}{ll}
      1, & \hbox{if $h(x_i)>h(x_j)$;} \\
      0.5, & \hbox{if $h(x_i)=h(x_j)$;} \\
      0, & \hbox{if $h(x_i)<h(x_j)$.}
    \end{array}
  \right.
  \label{ds}
\end{equation}
For more details of MAUC, readers are referred to \cite{Hand2001}. Moreover, Hand and Till have proved that MAUC measures the performance of classifiers independent of cost.

\section{Post Processing Classifiers}
Since a base classifier measured by the MAUC metric outputs a numerical score matrix, it must be converted to a discrete classifier that outputs class labels for practical use with the objective that the total cost is minimized. As mentioned above, there are mainly two categories of post-processing methods to implement the conversion.
\subsection{Re-optimization}
Lachiche and Flach \cite{Lachiche2003} have formulated this conversion problem as an optimization problem. Formally, given a $c\times{c}$ cost matrix $Cost$ ($Cost(i, j)$ is the cost of misclassifying an instance from class \emph{i} into class \emph{j}) and a $n\times{c}$ output matrix $M$ ($M_{ij}$ is the confidence that instance \emph{i} belongs to class \emph{j}), the problem is finding a weight vector ($w_1, w_2, \ldots, w_c$) to minimize
\begin{equation}
  \sum_{i=1}^{n}{Cost\big(y_i, \argmax_{j}\{w_jM_{ij}\}\big)}.
  \label{weighting}
\end{equation}
Furthermore, \cite{Bourke2008} has proved that the decision version of this formulated problem is NP-complete. Therefore, only approximation-type algorithms could be used in practice. In the next, we will briefly review three re-optimization methods.
\subsubsection{LF Method}
This method was proposed in \cite{Lachiche2003}. As a greedy hill-climbing search method, the LF method starts the search by assuming $w_1=1$. Then, in step \emph{i}, the LF considers instances of class \emph{i} against instances of classes whose weights have been fixed, and use the threshold moving method to find the weight $w_i$. This procedure is repeated until all the weights have been set.
\subsubsection{MetaClass Method}
Similar to the LF method, the MetaClass method proposed in \cite{Bourke2008} also reduces the multi-class optimization problem into a batch of binary optimization problems. Specifically, the output of MetaClass would be a tree. At the bottom of the tree, all of the classes will be split up into two groups, with approximately equal number of instances in each group. Then the threshold moving approach can be applied by treating these two groups as meta classes to form a binary problem. This procedure is recurred until each group contains only one class.
\subsubsection{Re-optimization by Genetic Algorithm}
In \cite{Bourke2008}, the authors also use the genetic algorithm (GA) to find the optimal weight vector directly. Upon the termination of GA, a direct pattern search was also employed to further tune the results.

In the experimental studies conducted in \cite{Bourke2008} on 24 UCI data sets \cite{Asuncion2007} using Naive Bayes as the base classification algorithm. These three re-optimization methods have shown advantages over several other re-optimization methods.

\subsection{Calibration}
Instead of formulating the conversion problem as Eq. (\ref{weighting}), one can also calibrate the output matrix into posterior probabilities, and make optimal decisions according to the Baysian rule,
\begin{equation}
    \argmin_{i}{\sum_{j=1}^{c}p(j)Cost(i, j)}.
\end{equation}
where $p(i)$ is the prior probability of class $i$. Two popular and widely used calibration methods are the Platt method \cite{Platt2000}, and the isotonic regression \cite{Robertson_Wright_Dykstra_1988} method recommend in \cite{Zadrozny2002}.
\subsubsection{Platt Method}
The Platt calibration method \cite{Platt2000} was first proposed for calibrating the output of SVM into probabilities and was used on other classification algorithms afterwards. Assuming that the output of classifier $H$ for instance $x_i$ is $H_i$ indicating that the confidence that instance $x_i$ belongs to class 1. The calibrated probability for this instance is,
\begin{equation}
  p_i=\frac{1}{1+\text{exp}(AH_i+B)}
\end{equation}
$A$ and $B$ are the parameters that learned from the data set to minimize the following target.
\begin{equation}
  \min F(A, B) = -\sum_{i=1}^{n}{\big(t_i\log(p_i)+(1-t_i)\log(1-p_i)\big)}
\end{equation}
where
\[
  t_i=
  \left\{
    \begin{array}{ll}
      \frac{N_++1}{N_++2}, & \hbox{if $y_i=1$;} \\
      \frac{1}{N_-+2}, & \hbox{if $y_i=-1$.}
    \end{array}
  \right.
  \label{s}
\]
In particular, Platt used a Levenberg-Marquardt (LM) algorithm \cite{Press1992} to find the best configuration of $A$ and $B$. Furthermore, Lin et at. \cite{Lin_Lin_Weng_2007} proved that this parameter tuning problem is in fact a convex optimization problem, and proposed a more robust and theoretically converging algorithm.
\subsubsection{Isotonic Regression}
Isotonic regression \cite{Robertson_Wright_Dykstra_1988} is a non-parametric form of regression which only assumes that the classifier ranks the instances correctly according to their posterior probabilities. The most widely used method for solving this regression problem is the Pool Adjacent Violators (PAV) method \cite{Ayer_Brunk_Ewing_Reid_Silverman_1955}. In \cite{Zadrozny2002}, the authors proposed using the PAV method to convert the output scores into posterior probabilities.  Briefly speaking, the PAV method firstly sorts the scores, then initializes the weight of each instance with its class label (0 or 1). Afterwards, the method finds a pair of instances that the weights of them are in reverse order as the scores (an adjacent violator), and replaces the weights of them with their mean weight. This procedure is repeated until there is no violator anymore. The weights of instances are their final posterior probabilities.

Platt method is a parametric method which performs well when its parametrical assumption is validated. On the other hand, the PAV method is a non-parametric method which counts more heavily on the training data. Note that both of these two calibration methods are designed for binary problems. In \cite{Zadrozny2002}, the authors have compared several approaches for combining binary sub-problems' calibrated posterior probabilities into multi-class form, and the simple normalization method was recommend. Hence, we will use the normalization method throughout this work.

\section{Experiments}
\subsection{Data Sets}
 Twenty-six multi-class data sets from UCI repository \cite{Asuncion2007} were used in our experiments. The detailed information of these data sets are presented in Table \ref{data}. These data sets are of virous characters, some of these data sets are imbalanced (e.g., the arrhythmia data set), the others are quite balanced (e.g., the synthetic data set).  The number of features ranges from 4 to 649, and the number of classes ranges from 3 to 26.
\begin{table*}
\caption{Summary of the data sets used in the experiments.}
\centering
\newsavebox{\databox}
\begin{lrbox}{\databox}
\begin{tabular}{l*{3}{c}}
\toprule
\textbf{Data Set} & \textbf{No. of Features} & \textbf{No. of Classes} & \textbf{Class Distribution}\\
\midrule
arrhythmia & 279 & 7 & 245, 44, 15, 15, 25, 50, 22\\
balance-scale & 4 & 3 & 49, 288, 288\\
car & 6 & 4 & 1210, 384, 65, 69\\
chess & 6 & 18 & 2796, 27, 78, 246, 81, 198, 471, 592, 683, 1433, 1712, 1985, 2854, 3597, 4194, 4553, 2166, 390\\
contraceptive & 9 & 3 & 629, 333, 511\\
dermatology & 34 & 6 & 112, 61, 72, 49, 52, 20\\
ecoli & 7 & 5 & 143, 77, 35, 20, 52\\
glass & 9 & 4 & 70, 76, 17, 29\\
hayes-roth & 3 & 3 & 65, 64, 31\\
leatter-recognition & 16 & 26 & 789, 766, 736, 805, 768, 775, 773, 734, 755, 747, 739, 761, 792, 783, 753, 803, 783, 758, 748, 796, 813, 764, 752, 787, 786, 734\\
mfeat & 649 & 10 & 200, 200, 200, 200, 200, 200, 200, 200, 200, 200\\
new-thyroid & 5 & 3 & 150, 35, 30\\
nursery & 8 & 4 & 4266, 4320, 328, 4044\\
optdigits & 64 & 10 & 554, 571, 557, 572, 568, 558, 558, 566, 554, 562\\
page-blocks & 10 & 5 & 4913, 329, 28, 88, 115\\
pendigits & 16 & 10 & 1143, 1143, 1144, 1055, 1144, 1055, 1056, 1142, 1055, 1055\\
satellite & 36 & 6 & 1533, 703, 1358, 626, 707, 1508\\
segment & 19 & 7 & 330, 330, 330, 330, 330, 330, 330\\
splice & 60 & 3 & 767, 768, 1655\\
synthetic & 60 & 6 & 100, 100, 100, 100, 100, 100\\
thyroid-allhypo & 28 & 3 & 3481, 194, 95\\
thyroid-allrep & 28 & 4 & 3648, 38, 52, 34\\
thyroid-ann & 21 & 3 & 166, 368, 6666\\
waveform21 & 21 & 3 & 1657, 1647, 1696\\
waveform40 & 40 & 3 & 1692, 1653, 1655\\
xa & 18 & 4 & 212, 217, 218, 199\\
\bottomrule
\end{tabular}
\end{lrbox}
\scalebox{0.83}{\usebox{\databox}}
\label{data}
\end{table*}

\subsection{Experimental Design}
In the experiments, we used the MAUC decomposition based Classification (MDC) \cite{Tang2011}, C4.5, Naive Bayes (NB), Artificial Neural Networks (ANN) and Support Vector Machine (SVM) as the base classifiers, and all the 5 conversion methods listed in Section \uppercase\expandafter{\romannumeral3} were employed to convert the base classifiers into discrete classifiers. MDC is a newly proposed method for obtaining large MAUC directly. It breaks the MAUC into a batch of components, and then maximizes each component independently. The experiments conducted in \cite{Tang2011} showed the capability of MDC in obtaining large MAUC. All the other base classifiers are widely used in the literature. To establish a baseline as well as to check whether post processing classifiers can really reduce the total cost, we also used the dummy post-processing method that assigns an instance to the class with the largest score, namely the \emph{Raw} method. For each of the 26 data sets, 20 different runs of 5-fold cross-validation were executed. With the testing output matrix of each base classifier, we can first measure the MAUC and accuracy of each classification method on every data set. Afterwards, 20 cost matrices were randomly generated for each data set, one matrix for one run of cross-validation. Next, the post-processing methods were applied on the training output matrices to find the best conversion, and then the testing output matrices were used to evaluate the total misclassification cost.

In specific, the cost matrices were generated analogous to the approach used in \cite{Domingos_1999}. That is, the $Cost(i, j) (i\neq{j})$ was sampled with uniform probability from the interval [$0, 2000p(i)/p(j)$], where $p(i)$ and $p(j)$ are the prior probabilities of class $i$ and class $j$ respectively. Besides, without loss of generality,  $Cost(i, i)$ was set to $0$.

All of the classification algorithms and post-processing methods were implemented or directly used in the WEKA platform \cite{Witten2000} except that the GA re-optimization method was employed from the genetic algorithm and direct search toolbox \cite{GAinMatlab} of Matlab directly. The parameters of post-processing methods and MDC were set according to corresponding paper while the search range of GA is set to $[-1,1]$ for each element of the weight vector. All the other parameters were kept as default values in WEKA.

\subsection{Results}
Table \ref{maucs} presents the testing MAUC and accuracy of base classifiers on the 26 data sets. Each value is averaged over 20 different runs of 5-fold cross-validation and is rounded to 0.001. The best MAUC and the best accuracy for each data set are in boldface. Tables \ref{resA} and \ref{resB} present the results of the average misclassification costs per instance in testing. The LF, MetaClass, and GA re-optimzation methods are tabulated in Table \ref{resA}, the Platt, PAV and the Raw method are tabulated in Table \ref{resB}. Each element in these tables is averaged over 20 runs of 5-fold cross-validation. The smallest cost on each data set is also in boldface.

To see whether the post-processing methods can really reduce the total cost, the average percentage of cost reduction by each method relative to the Raw method is summarized in Table \ref{lifting}. If the post-processing methods can indeed reduce the cost on average, the number would be positive, otherwise it would be negative. As we can see clearly that the LF and MetaClass methods actually \emph{cannot} reduce the testing cost on average while GA, Platt and PAV methods can (except PAV was applied onto the MDC base classifier). Furthermore, the numbers of win, draw and loss compared to the Raw method  in testing cost are also included in the parenthesis. For example, we can see that on average the PAV method increased the cost by $10.77\%$ relative to the Raw method on the MDC base classifier. However, the win-draw-loss counting indicates that PAV actually reduces the cost on 22 out of 26 data sets, only increases the cost on the other 4 data sets. In summary, we can see that  the Platt method can reduce the cost maximally when MDC, ANN or SVM was used as the base classifier, and the PAV performs the best with C4.5 and NB.

Although GA can reduce the testing cost, it was always outperformed by the Platt or PAV method (there is no boldfaced number in Table \ref{resA}). Besides, the GA method is much more time consuming than the other methods. Between the Platt and PAV method, the former performed better. Through a closer look at the total cost obtained on training data, we found that the PAV method achieved superior performance. This observation suggests that the PAV method is prone to over-fitting. For the sake of clarity, the results on training data are not presented in this paper, but are made available at request. Given the above comparisons, we may draw the conclusion that simple calibration methods (such as the Platt and PAV) that convert the output matrix into posterior probabilities perform better than existing sophisticated post re-optimization methods.

\begin{table*}[!htbp]
\caption{The MAUC and Accuracy of Base Classifiers. The Best MAUC and Accuracy on Each Data Set are in Boldface.}
\centering
\begin{tabular}{l|*{5}{c}|*{5}{c}}
\toprule
& \multicolumn{5}{c}{MAUC} & \multicolumn{5}{c}{Accuracy}\\
& MDC & C4.5 & NB & ANN & SVM & MDC & C4.5 & NB & ANN & SVM\\
\midrule
arrhythmia &$\textbf{0.856}$ &$0.766$ &$0.810$ &$0.839$ &$0.481$ &$0.615$ &$0.709$ &$0.645$ &$\textbf{0.728}$ &$0.589$\\
balance-scale &$0.740$ &$0.705$ &$0.887$ &$\textbf{0.944}$ &$0.940$ &$\textbf{0.922}$ &$0.777$ &$0.899$ &$0.911$ &$0.889$\\
car &$0.963$ &$0.974$ &$0.898$ &$0.930$ &$\textbf{0.984}$ &$0.071$ &$\textbf{0.955}$ &$0.763$ &$0.884$ &$0.943$\\
chess &$0.871$ &$0.950$ &$0.838$ &$0.834$ &$\textbf{0.969}$ &$0.179$ &$\textbf{0.769}$ &$0.303$ &$0.440$ &$0.674$\\
contraceptive &$\textbf{0.730}$ &$0.662$ &$0.687$ &$0.714$ &$0.718$ &$0.364$ &$0.524$ &$0.489$ &$0.543$ &$\textbf{0.555}$\\
dermatology &$0.996$ &$0.975$ &$\textbf{0.997}$ &$0.996$ &$0.990$ &$0.944$ &$0.958$ &$\textbf{0.973}$ &$0.969$ &$0.930$\\
ecoli &$0.953$ &$0.891$ &$0.965$ &$0.951$ &$\textbf{0.966}$ &$0.405$ &$0.835$ &$\textbf{0.868}$ &$0.861$ &$0.859$\\
glass &$\textbf{0.883}$ &$0.796$ &$0.811$ &$0.828$ &$0.853$ &$0.493$ &$0.703$ &$0.526$ &$0.693$ &$\textbf{0.715}$\\
hayes-roth &$\textbf{0.972}$ &$0.951$ &$0.910$ &$0.837$ &$0.966$ &$0.600$ &$0.823$ &$0.728$ &$0.711$ &$\textbf{0.827}$\\
letter-recognition &$0.986$ &$0.951$ &$0.957$ &$0.959$ &$\textbf{1.000}$ &$0.310$ &$0.874$ &$0.641$ &$0.825$ &$\textbf{0.974}$\\
mfeat &$\textbf{0.998}$ &$0.972$ &$0.990$ &$0.998$ &$0.494$ &$0.964$ &$0.943$ &$0.951$ &$\textbf{0.984}$ &$0.099$\\
new-thyroid &$0.996$ &$0.936$ &$0.996$ &$\textbf{0.997}$ &$0.970$ &$0.922$ &$0.924$ &$\textbf{0.969}$ &$0.913$ &$0.877$\\
nursery &$0.949$ &$0.999$ &$0.964$ &$0.984$ &$\textbf{0.999}$ &$0.645$ &$\textbf{0.994}$ &$0.901$ &$0.948$ &$0.986$\\
optdigits &$0.997$ &$0.952$ &$0.986$ &$0.999$ &$\textbf{0.999}$ &$0.628$ &$0.902$ &$0.913$ &$\textbf{0.983}$ &$0.827$\\
page-blocks &$\textbf{0.988}$ &$0.935$ &$0.951$ &$0.922$ &$0.719$ &$0.682$ &$\textbf{0.970}$ &$0.895$ &$0.960$ &$0.913$\\
pendigits &$\textbf{0.997}$ &$0.983$ &$0.979$ &$0.967$ &$0.386$ &$0.629$ &$\textbf{0.962}$ &$0.858$ &$0.946$ &$0.122$\\
satellite &$\textbf{0.981}$ &$0.915$ &$0.956$ &$0.978$ &$0.703$ &$0.765$ &$0.862$ &$0.796$ &$\textbf{0.893}$ &$0.298$\\
segment &$\textbf{0.997}$ &$0.985$ &$0.970$ &$0.996$ &$0.991$ &$0.532$ &$\textbf{0.965}$ &$0.801$ &$0.961$ &$0.675$\\
splice &$\textbf{0.989}$ &$0.942$ &$0.975$ &$0.956$ &$0.966$ &$0.489$ &$\textbf{0.925}$ &$0.887$ &$0.859$ &$0.869$\\
synthetic &$0.999$ &$0.962$ &$0.998$ &$\textbf{1.000}$ &$0.236$ &$0.635$ &$0.919$ &$0.945$ &$\textbf{0.993}$ &$0.166$\\
thyroid-allhypo &$\textbf{0.999}$ &$0.990$ &$0.927$ &$0.909$ &$0.893$ &$0.199$ &$\textbf{0.996}$ &$0.954$ &$0.952$ &$0.932$\\
thyroid-allrep &$\textbf{0.961}$ &$0.916$ &$0.916$ &$0.907$ &$0.734$ &$0.025$ &$\textbf{0.991}$ &$0.930$ &$0.967$ &$0.966$\\
thyroid-ann &$\textbf{1.000}$ &$0.993$ &$0.942$ &$0.952$ &$0.962$ &$0.173$ &$\textbf{0.996}$ &$0.954$ &$0.961$ &$0.963$\\
waveform21 &$0.964$ &$0.848$ &$0.959$ &$0.965$ &$\textbf{0.965}$ &$0.830$ &$0.764$ &$0.809$ &$0.841$ &$\textbf{0.860}$\\
waveform40 &$0.964$ &$0.828$ &$0.956$ &$0.962$ &$\textbf{0.969}$ &$0.852$ &$0.752$ &$0.800$ &$0.835$ &$\textbf{0.863}$\\
xa &$0.920$ &$0.855$ &$0.772$ &$\textbf{0.947}$ &$0.529$ &$0.495$ &$0.723$ &$0.450$ &$\textbf{0.812}$ &$0.261$\\
\bottomrule
\end{tabular}
\label{maucs}
\end{table*}
To examine the correlation between the MAUC and total cost of classifiers. We need to specify the post-processing method. Particularly, only GA, Platt and PAV methods were included in this analysis, while the LF and MetaClass methods were abandoned for their poor performance showed above. In fact, if the post-processing methods can not reduce the total cost comparing to the Raw method, the relationship between the MAUC and total cost established on this method is meaningless.

Table \ref{rcc} presents the Spearman correlation coefficient \cite{Myers_Well_2003} between the MAUC/accuracy of classifiers and their total cost combining with the GA, Platt or PAV method. Unlike the Person correlation coefficient, Spearman correlation coefficient measures the correlation between two variables based on their ranks instead of values. Therefore, it suits the aim of the experiments quite well. Generally, we can see that the average Spearman correlation coefficient between the classifiers' MAUC and their total cost is $-0.45$ if the base classifiers are post processed by the GA method, and the coefficients for Platt and PAV are $-0.80$ and $-0.54$. On the other hand, these values for accuracy are $-0.53$, $-0.31$ and $-0.47$.

We furthermore conducted statistical tests \cite{Demsar2006} to compare the difference among these correlations. Since we have the results of multiple methods (\{MAUC, Accuracy\}$\times$\{GA, Platt, PAV\}) on multiple data sets, and no particular method should be considered as the control method in advance. Hence, the Friedman test was used firstly to detect whether there are differences among these methods, then the Holm test was used to further find which pairs of methods are significantly different. The significance level was set to $5\%$, and the test procedures were implemented by \cite{Garcia_Herrera_2008}. The results are summarized in Table \ref{testtable}. We can see that the MAUC-Platt obtained the highest ranking, which indicates the relationship between MAUC and the total cost based on Platt is the strongest. Moreover, the P-value of Friedman test is at the level of 1E-5, so it is almost sure that there are indeed differences among these methods. Then, the Holm test reported that the first four pairs of comparisons are statistically significant. For instance, the first pair of comparison indicates that the correlation between MAUC and total cost based on Platt is significantly stronger than that between the accuracy and total cost based on the Platt. Hence, MAUC is more useful than accuracy when using Platt as post-processing method. In addition, if we focus on the smallest cost obtained on each data set, we can see that on 22 out of 26 data sets, the smallest costs were obtained on the classifiers with the largest MAUC. In comparison, the largest accuracy led to the smallest costs on only 10 data sets.
\begin{table*}[!htbp]
\caption{Comparison of the testing cost per instance on each data set (for LF, MetaClass, GA methods). The Smallest Cost on Each Data Set is in Boldface.}
\centering
\newsavebox{\resultboxA}
\begin{lrbox}{\resultboxA}
\begin{tabular}{l|*{5}{c}|*{5}{c}|*{5}{c}}
\toprule
& \multicolumn{5}{c}{LF} & \multicolumn{5}{c}{MetaClass} & \multicolumn{5}{c}{GA}\\
& MDC & C4.5 & NB & ANN & SVM & MDC & C4.5 & NB & ANN & SVM & MDC & C4.5 & NB & ANN & SVM\\
\midrule
arrhythmia & $449.71$  & $2415.87$  & $1794.93$  & $3005.16$  & $3619.48$  & $198.33$  & $386.84$  & $348.86$  & $221.44$  & $218.25$  & $505.15$  & $414.99$  & $278.42$  & $656.48$  & $298.09$ \\
balance-scale & $573.36$  & $829.84$  & $929.7$  & $829.29$  & $686.22$  & $239.67$  & $295.98$  & $173.85$  & $204.54$  & $171.09$  & $221.44$  & $312.65$  & $96.18$  & $91.16$  & $118.34$ \\
car & $539.33$  & $727.74$  & $2227.87$  & $1623.77$  & $882.76$  & $127.79$  & $394.5$  & $137.91$  & $158.21$  & $132.73$  & $68.19$  & $29.31$  & $70.95$  & $62.47$  & $46.73$ \\
chess & $43.22$  & $1338.71$  & $2229.31$  & $1582.94$  & $1771.64$  & $22.39$  & $1346.91$  & $18.09$  & $19.01$  & $23.83$  & $20.54$  & $229.85$  & $15.5$  & $15.47$  & $197.88$ \\
contraceptive & $854.83$  & $739.76$  & $859.02$  & $840.4$  & $834.19$  & $416.65$  & $458.26$  & $421.93$  & $437.37$  & $426.63$  & $366.58$  & $402.01$  & $360.59$  & $348.84$  & $347.56$ \\
dermatology & $589.54$  & $640.1$  & $937.67$  & $477.98$  & $296.12$  & $400.72$  & $562.72$  & $301.35$  & $317.89$  & $321.61$  & $43.09$  & $41.28$  & $18.26$  & $30.86$  & $68.12$ \\
ecoli & $525.39$  & $864.51$  & $578.94$  & $929.69$  & $550.14$  & $318.91$  & $666.45$  & $287.45$  & $294.47$  & $320.72$  & $215.63$  & $215.03$  & $140.23$  & $162.57$  & $137.18$ \\
glass & $532.15$  & $891.95$  & $1220.26$  & $1061.62$  & $1061.96$  & $313.8$  & $520.15$  & $295.35$  & $301.23$  & $310.8$  & $369.5$  & $358.56$  & $248.32$  & $306.03$  & $182.92$ \\
hayes-roth & $794.95$  & $322.66$  & $659.31$  & $722.29$  & $464.14$  & $377.25$  & $452.53$  & $471.78$  & $445.57$  & $456.37$  & $221.66$  & $166.9$  & $142.01$  & $254.21$  & $144.46$ \\
letter-recognition & $969.59$  & $833.88$  & $1005.84$  & $951.67$  & $491.55$  & $921.61$  & $947.21$  & $932.78$  & $931.66$  & $933.28$  & $285.35$  & $123.61$  & $309.37$  & $174.94$  & $27.99$ \\
mfeat & $310.3$  & $561.19$  & $691.15$  & $463.75$  & $1037.44$  & $845.1$  & $865.77$  & $867.0$  & $848.42$  & $874.38$  & $28.19$  & $53.46$  & $48.54$  & $18.69$  & $753.33$ \\
new-thyroid & $461.46$  & $819.05$  & $863.7$  & $1361.54$  & $251.88$  & $411.02$  & $413.53$  & $440.25$  & $410.21$  & $340.88$  & $233.38$  & $187.51$  & $77.2$  & $86.1$  & $259.5$ \\
nursery & $82.92$  & $335.42$  & $805.26$  & $691.35$  & $449.82$  & $314.57$  & $617.89$  & $157.8$  & $134.24$  & $122.22$  & $55.46$  & $6.35$  & $65.74$  & $41.51$  & $15.68$ \\
optdigits & $785.94$  & $694.7$  & $980.47$  & $729.84$  & $928.56$  & $867.43$  & $888.31$  & $893.2$  & $887.74$  & $889.02$  & $44.74$  & $95.76$  & $81.1$  & $20.66$  & $210.96$ \\
page-blocks & $112.36$  & $2689.61$  & $2318.3$  & $3387.31$  & $3127.77$  & $162.18$  & $223.13$  & $196.49$  & $72.22$  & $71.68$  & $32.92$  & $208.33$  & $17.2$  & $16.75$  & $1207.86$ \\
pendigits & $820.04$  & $557.06$  & $812.87$  & $790.57$  & $1082.11$  & $804.11$  & $899.96$  & $825.3$  & $824.42$  & $827.4$  & $142.0$  & $37.23$  & $130.68$  & $49.92$  & $714.17$ \\
satellite & $702.62$  & $630.61$  & $757.55$  & $734.83$  & $1268.15$  & $539.05$  & $639.78$  & $546.31$  & $517.52$  & $541.84$  & $110.26$  & $169.45$  & $192.12$  & $132.28$  & $632.09$ \\
segment & $751.95$  & $441.13$  & $792.4$  & $602.34$  & $781.91$  & $786.59$  & $847.74$  & $855.72$  & $838.92$  & $842.57$  & $111.84$  & $34.44$  & $124.04$  & $34.3$  & $377.29$ \\
splice & $367.3$  & $223.25$  & $384.71$  & $324.99$  & $463.35$  & $465.69$  & $545.98$  & $528.41$  & $530.56$  & $526.44$  & $222.25$  & $87.24$  & $97.93$  & $165.46$  & $132.37$ \\
synthetic & $726.9$  & $512.3$  & $672.0$  & $315.84$  & $1017.79$  & $651.07$  & $768.04$  & $775.92$  & $782.22$  & $783.37$  & $187.09$  & $84.02$  & $49.61$  & $11.13$  & $651.81$ \\
thyroid-allhypo & $837.83$  & $810.67$  & $1502.83$  & $1878.14$  & $482.83$  & $92.13$  & $114.31$  & $151.13$  & $135.27$  & $111.94$  & $9.99$  & $6.2$  & $39.49$  & $51.35$  & $1392.11$ \\
thyroid-allrep & $38.96$  & $1345.91$  & $2738.33$  & $2736.87$  & $2733.29$  & $40.41$  & $34.69$  & $34.96$  & $31.72$  & $30.4$  & $14.17$  & $107.67$  & $55.48$  & $83.33$  & $2324.06$ \\
thyroid-ann & $99.66$  & $839.39$  & $474.34$  & $1214.26$  & $565.18$  & $82.34$  & $98.62$  & $98.78$  & $99.21$  & $98.79$  & $5.01$  & $24.87$  & $36.83$  & $41.76$  & $28.09$ \\
waveform21 & $483.83$  & $347.57$  & $353.47$  & $212.76$  & $231.67$  & $438.98$  & $575.68$  & $566.07$  & $577.7$  & $573.18$  & $137.09$  & $236.24$  & $138.73$  & $154.75$  & $133.28$ \\
waveform40 & $391.99$  & $334.58$  & $296.51$  & $210.89$  & $219.0$  & $477.73$  & $561.83$  & $634.42$  & $621.41$  & $619.14$  & $136.72$  & $248.37$  & $147.14$  & $168.38$  & $136.7$ \\
xa & $733.82$  & $459.42$  & $801.31$  & $461.82$  & $941.3$  & $492.55$  & $624.2$  & $492.02$  & $491.0$  & $679.1$  & $280.69$  & $256.49$  & $347.93$  & $172.38$  & $521.49$ \\
\bottomrule
\end{tabular}
\end{lrbox}
\scalebox{0.87}{\usebox{\resultboxA}}
\label{resA}
\end{table*}
\begin{table*}[!htbp]
\caption{Comparison of the testing cost per instance on each data set (for Platt, PAV, Raw methods). The Smallest Cost on Each Data Set is in Boldface.}
\centering
\newsavebox{\resultboxB}
\begin{lrbox}{\resultboxB}
\begin{tabular}{l|*{5}{c}|*{5}{c}|*{5}{c}}
\toprule
& \multicolumn{5}{c}{Platt} & \multicolumn{5}{c}{PAV} & \multicolumn{5}{c}{Raw}\\
& MDC & C4.5 & NB & ANN & SVM & MDC & C4.5 & NB & ANN & SVM & MDC & C4.5 & NB & ANN & SVM\\
\midrule
arrhythmia & $\textbf{120.57}$  & $153.28$  & $167.58$  & $153.5$  & $183.06$  & $1494.19$  & $347.75$  & $229.53$  & $417.5$  & $180.85$  & $1638.88$  & $1118.2$  & $1076.15$  & $1278.99$  & $3619.48$ \\
balance-scale & $138.4$  & $184.93$  & $104.65$  & $84.16$  & $78.65$  & $191.14$  & $300.75$  & $99.09$  & $\textbf{75.87}$  & $123.38$  & $518.36$  & $630.06$  & $568.65$  & $227.22$  & $467.09$ \\
car & $47.36$  & $57.65$  & $61.7$  & $73.07$  & $46.1$  & $31.45$  & $\textbf{24.98}$  & $44.32$  & $48.91$  & $29.94$  & $128.24$  & $84.97$  & $956.66$  & $189.72$  & $108.35$ \\
chess & $15.26$  & $14.13$  & $16.06$  & $16.0$  & $14.56$  & $15.51$  & $71.34$  & $15.37$  & $15.25$  & $\textbf{14.09}$  & $549.58$  & $323.79$  & $1105.5$  & $942.67$  & $454.47$ \\
contraceptive & $\textbf{319.8}$  & $384.99$  & $351.08$  & $336.04$  & $338.95$  & $322.32$  & $391.12$  & $352.8$  & $334.48$  & $339.44$  & $508.36$  & $540.09$  & $507.89$  & $502.16$  & $551.97$ \\
dermatology & $24.09$  & $43.64$  & $19.45$  & $23.95$  & $51.24$  & $70.19$  & $41.18$  & $\textbf{17.06}$  & $24.8$  & $67.22$  & $62.79$  & $41.78$  & $19.07$  & $24.52$  & $70.69$ \\
ecoli & $126.28$  & $164.11$  & $110.8$  & $133.84$  & $\textbf{106.66}$  & $312.22$  & $182.96$  & $110.36$  & $134.23$  & $106.68$  & $233.94$  & $278.31$  & $152.2$  & $182.63$  & $190.65$ \\
glass & $199.31$  & $245.27$  & $206.65$  & $203.27$  & $\textbf{162.08}$  & $458.14$  & $299.86$  & $219.99$  & $255.95$  & $188.01$  & $600.27$  & $516.92$  & $610.68$  & $626.84$  & $621.81$ \\
hayes-roth & $\textbf{88.38}$  & $145.73$  & $177.26$  & $240.09$  & $130.81$  & $88.53$  & $134.71$  & $147.02$  & $241.36$  & $113.89$  & $396.78$  & $195.85$  & $302.31$  & $339.97$  & $186.79$ \\
letter-recognition & $146.79$  & $119.31$  & $285.78$  & $159.69$  & $\textbf{21.83}$  & $145.31$  & $113.98$  & $252.27$  & $152.75$  & $23.48$  & $670.39$  & $126.26$  & $361.26$  & $176.66$  & $25.85$ \\
mfeat & $\textbf{11.5}$  & $52.86$  & $48.55$  & $14.55$  & $731.5$  & $32.48$  & $52.19$  & $40.78$  & $17.4$  & $647.93$  & $33.59$  & $54.51$  & $48.69$  & $15.41$  & $894.41$ \\
new-thyroid & $48.66$  & $143.2$  & $85.12$  & $\textbf{44.6}$  & $115.74$  & $187.61$  & $182.35$  & $56.96$  & $89.4$  & $955.91$  & $407.62$  & $230.45$  & $138.16$  & $407.26$  & $148.15$ \\
nursery & $39.53$  & $25.55$  & $43.77$  & $40.19$  & $10.63$  & $34.63$  & $\textbf{3.7}$  & $37.15$  & $23.02$  & $7.92$  & $701.92$  & $10.9$  & $401.06$  & $370.08$  & $35.32$ \\
optdigits & $32.5$  & $95.4$  & $80.7$  & $\textbf{16.17}$  & $133.82$  & $39.67$  & $93.01$  & $71.2$  & $21.18$  & $328.27$  & $362.12$  & $97.72$  & $83.6$  & $16.93$  & $163.46$ \\
page-blocks & $\textbf{6.83}$  & $9.18$  & $13.71$  & $12.56$  & $20.31$  & $346.32$  & $168.79$  & $10.29$  & $11.53$  & $391.93$  & $387.67$  & $628.01$  & $877.49$  & $1269.78$  & $2825.05$ \\
pendigits & $\textbf{31.54}$  & $37.15$  & $135.57$  & $48.85$  & $795.62$  & $32.84$  & $36.2$  & $116.38$  & $45.32$  & $611.45$  & $383.42$  & $37.71$  & $146.97$  & $56.12$  & $934.03$ \\
satellite & $\textbf{88.93}$  & $166.21$  & $172.93$  & $115.51$  & $563.04$  & $93.69$  & $161.43$  & $146.59$  & $105.08$  & $442.24$  & $306.9$  & $190.01$  & $222.87$  & $149.61$  & $1115.04$ \\
segment & $\textbf{25.0}$  & $34.55$  & $158.4$  & $32.84$  & $253.85$  & $27.98$  & $33.23$  & $100.81$  & $31.22$  & $501.74$  & $476.25$  & $35.03$  & $212.71$  & $39.01$  & $314.29$ \\
splice & $41.75$  & $90.05$  & $105.51$  & $156.96$  & $132.13$  & $\textbf{40.84}$  & $86.33$  & $92.28$  & $168.06$  & $122.56$  & $261.58$  & $94.85$  & $131.76$  & $177.8$  & $172.91$ \\
synthetic & $23.45$  & $84.47$  & $58.29$  & $\textbf{6.26}$  & $545.57$  & $47.25$  & $83.2$  & $39.46$  & $7.39$  & $545.57$  & $421.19$  & $85.05$  & $62.85$  & $7.28$  & $881.53$ \\
thyroid-allhypo & $\textbf{1.64}$  & $19.57$  & $35.55$  & $36.0$  & $112.8$  & $95.09$  & $6.2$  & $37.56$  & $63.37$  & $1456.11$  & $68.87$  & $29.81$  & $657.4$  & $697.9$  & $1424.76$ \\
thyroid-allrep & $\textbf{7.17}$  & $12.03$  & $14.81$  & $15.81$  & $24.18$  & $545.57$  & $83.87$  & $149.25$  & $100.42$  & $2555.89$  & $34.18$  & $207.1$  & $870.09$  & $1206.05$  & $2661.89$ \\
thyroid-ann & $\textbf{0.93}$  & $21.46$  & $34.26$  & $31.35$  & $32.01$  & $31.68$  & $24.87$  & $29.81$  & $32.16$  & $32.94$  & $82.64$  & $33.19$  & $704.03$  & $535.93$  & $567.74$ \\
waveform21 & $126.6$  & $235.56$  & $156.28$  & $152.88$  & $129.51$  & $127.04$  & $232.28$  & $131.57$  & $145.37$  & $\textbf{124.4}$  & $176.53$  & $244.79$  & $200.73$  & $165.7$  & $145.71$ \\
waveform40 & $129.41$  & $251.02$  & $163.08$  & $165.35$  & $129.63$  & $129.62$  & $247.64$  & $143.81$  & $167.22$  & $\textbf{126.92}$  & $151.18$  & $252.2$  & $191.43$  & $168.05$  & $138.87$ \\
xa & $185.86$  & $253.3$  & $374.83$  & $167.82$  & $609.24$  & $204.43$  & $247.43$  & $330.85$  & $\textbf{160.66}$  & $472.05$  & $447.6$  & $279.86$  & $484.76$  & $192.83$  & $830.82$ \\
\bottomrule
\end{tabular}
\end{lrbox}
\scalebox{0.87}{\usebox{\resultboxB}}
\label{resB}
\end{table*}
\begin{table*}
\caption{Relative Testing Cost Reduction of Each Conversion Method Comparing to the Raw Method. Number of win-draw-lose Compared to the Raw Method is in the Parenthesis.}
\centering
\begin{tabular}{l*{5}{c}}
\toprule
& LF & MetaClass & GA & Platt & PAV\\
\midrule
MDC & $-157.26\% (5-0-21)$ & $-136.76\% (10-0-16)$ & $53.39\% (26-0-0)$ & $74.76\% (26-0-0)$ & $-10.77\% (22-0-4)$\\
C4.5 & $-686.15\% (0-0-26)$ & $-672.93\% (5-0-21)$ & $24.02\% (26-0-0)$ & $24.61\% (24-0-2)$ & $30.87\% (26-0-0)$\\
NB & $-451.68\% (1-0-25)$ & $-254.76\% (11-0-15)$ & $47.05\% (26-0-0)$ & $46.26\% (25-0-1)$ & $53.37\% (26-0-0)$\\
ANN & $-763.82\% (0-0-26)$ & $-1029.16\% (11-0-15)$ & $33.61\% (21-0-5)$ & $44.38\% (26-0-0)$ & $41.50\% (22-0-4)$\\
SVM & $-222.97\% (2-1-23)$ & $-197.76\% (14-0-12)$ & $27.37\% (22-0-4)$ & $48.64\% (26-0-0)$ & $13.59\% (22-0-4)$\\
\bottomrule
\end{tabular}
\label{lifting}
\end{table*}
\begin{table}
\caption{Spearman correlation coefficient between classifiers' MAUC/Accuracy and their total Cost.}
\centering
\newsavebox{\rccbox}
\begin{lrbox}{\rccbox}
\begin{tabular}{l|*{3}{r}|*{3}{r}}
\toprule
& \multicolumn{3}{c}{MAUC} & \multicolumn{3}{c}{Accuracy}\\
& GA & Platt & PAV & GA & Platt & PAV\\
\midrule
arrhythmia &$0.6$ &$-0.7$ &$0.9$ &$0.5$ &$-0.3$ &$0.4$\\
balance-scale &$-0.9$ &$-0.9$ &$-0.9$ &$-0.4$ &$-0.1$ &$-0.4$\\
car &$-0.8$ &$-0.8$ &$-0.8$ &$-0.9$ &$-0.1$ &$-0.6$\\
chess &$0.9$ &$-0.8$ &$0.0$ &$0.6$ &$-0.7$ &$0.0$\\
contraceptive &$-0.4$ &$-0.9$ &$-0.9$ &$-0.7$ &$0.2$ &$0.2$\\
dermatology &$-0.7$ &$-0.9$ &$-0.6$ &$-1.0$ &$-0.9$ &$-0.9$\\
ecoli &$-0.7$ &$-1.0$ &$-0.7$ &$-0.7$ &$-0.3$ &$-0.7$\\
glass &$0.1$ &$-0.9$ &$0.1$ &$-0.6$ &$-0.1$ &$-0.6$\\
hayes-roth &$-0.1$ &$-1.0$ &$-1.0$ &$-0.6$ &$0.0$ &$0.0$\\
letter-recognition &$-0.3$ &$-0.4$ &$-0.4$ &$-0.9$ &$-0.7$ &$-0.7$\\
mfeat &$-0.9$ &$-1.0$ &$-0.9$ &$-1.0$ &$-0.9$ &$-1.0$\\
new-thyroid &$-0.6$ &$-0.9$ &$-0.6$ &$-0.7$ &$0.2$ &$-0.7$\\
nursery &$-0.8$ &$-0.7$ &$-0.8$ &$-0.9$ &$-0.6$ &$-0.9$\\
optdigits &$0.0$ &$0.0$ &$0.0$ &$-0.4$ &$-0.4$ &$-0.4$\\
page-blocks &$-0.3$ &$-0.7$ &$-0.3$ &$0.1$ &$0.1$ &$-0.1$\\
pendigits &$-0.3$ &$-0.9$ &$-0.9$ &$-1.0$ &$-0.4$ &$-0.4$\\
satellite &$-0.9$ &$-0.9$ &$-1.0$ &$-0.4$ &$-0.4$ &$-0.3$\\
segment &$-0.3$ &$-0.7$ &$-0.7$ &$-0.6$ &$0.1$ &$0.1$\\
splice &$0.6$ &$-0.4$ &$-0.4$ &$-1.0$ &$0.0$ &$0.0$\\
synthetic &$-0.7$ &$-1.0$ &$-0.9$ &$-1.0$ &$-0.7$ &$-0.9$\\
thyroid-allhypo &$-0.9$ &$-1.0$ &$-0.4$ &$-0.4$ &$0.0$ &$-0.9$\\
thyroid-allrep &$-0.9$ &$-0.9$ &$-0.1$ &$0.7$ &$0.3$ &$-0.7$\\
thyroid-ann &$-0.9$ &$-0.9$ &$-0.1$ &$0.1$ &$0.1$ &$-0.1$\\
waveform21 &$-0.7$ &$-0.7$ &$-0.7$ &$-0.7$ &$-0.7$ &$-0.7$\\
waveform40 &$-0.9$ &$-0.8$ &$-0.9$ &$-0.9$ &$-0.8$ &$-0.9$\\
xa &$-0.9$ &$-1.0$ &$-1.0$ &$-1.0$ &$-0.9$ &$-0.9$\\
\midrule
Average &$-0.45$ &$-0.80$ &$-0.54$ &$-0.53$ &$-0.31$ &$-0.47$\\
\bottomrule
\end{tabular}
\end{lrbox}
\scalebox{0.9}{\usebox{\rccbox}}
\label{rcc}
\end{table}
\begin{table}[!htbp]
\caption{Statistical Test among the Spearman correlations, the significance level $\alpha=0.05$.}
\centering
\newsavebox{\testbox}
\begin{lrbox}{\testbox}
\begin{tabular}{l*{2}{c}}
\toprule
\multicolumn{3}{l}{\textbf{Friedman Test}}\\
\cmidrule(l){1-1}
Methods & Ranking & \\
\cmidrule(l){1-2}
Mauc-GA & $3.5769$ &\\
Mauc-Platt & $2.3846$ &\\
Mauc-PAV & $3.4038$&\\
Acc-GA & $2.7308$&\\
Acc-Platt & $4.9231$&\\
Acc-PAV & $3.9808$\\
\cmidrule(l){1-2}
\multicolumn{3}{l}{P-value computed by Friedman Test: 1.1697E-5}\\
\midrule
\midrule
\multicolumn{3}{l}{\textbf{Holm Test}}\\
\cmidrule(l){1-1}
Comparisons & P-value & $\alpha_{Holm}$\\
\midrule
MUAC-Platt vs.  Acc-Platt&9.9692E-7&0.0033\\
Acc-GA vs.  Acc-Platt&2.3881E-5&0.0036\\
MUAC-Platt vs.  Acc-PAV&0.0021&0.0038\\
MAUC-PAV vs.  Acc-Platt&0.0034&0.0042\\
MAUC-GA vs.  Acc-Platt&0.0095&0.0045\\
Acc-GA vs.  Acc-PAV&0.0160&0.0050\\
MAUC-GA vs.  MUAC-Platt&0.0216&0.0056\\
MUAC-Platt vs.  MAUC-PAV&0.0495&0.0063\\
Acc-Platt vs.  Acc-PAV&0.0694&0.0071\\
MAUC-GA vs.  Acc-GA&0.1029&0.0083\\
MAUC-PAV vs.  Acc-GA&0.1946&0.0100\\
MAUC-PAV vs.  Acc-PAV&0.2662&0.0125\\
MAUC-GA vs.  Acc-PAV&0.4364&0.0167\\
MUAC-Platt vs.  Acc-GA&0.5047&0.0250\\
MAUC-GA vs.  MAUC-PAV&0.7387&0.0500\\
\bottomrule
\end{tabular}
\end{lrbox}
\scalebox{0.93}{\usebox{\testbox}}
\label{testtable}
\end{table}

\section{Conclusions and Discussions}
Although MAUC can measure the performance of classifiers independent of cost, there is neither theoretical guarantees nor empirical studies published in the literature to justify the presumption that a larger MAUC of classifiers is more likely to lead to a smaller total cost. This paper has provided such an empirical justification.

To address this issue, one has to find a way to convert the base classifier that outputs a numerical matrix into a discrete classifier that outputs class labels. For binary problems, the conversion can be performed optimally and effectively by the simple threshold moving method. There is no such method for multi-class problems. We have employed two categories of methods to post process the base classifiers in this study. The first is re-optimization methods which formulate the conversion task as an optimization problem. The second is the calibration methods which transform the output scores into posterior probabilities and make decisions according to the Bayesian rule. Our comparative studies suggest that simple calibration methods outperform exiting sophisticated re-optimization methods.

With the MAUC values of base classifiers and the total cost of discrete classifiers, we analyzed the correlation between them. The Spearman correlation coefficients are negative when GA, Platt or PAV was used as the conversion method. This indicates that a larger MAUC is beneficial since it can indeed lead to a smaller total cost given the base classifier is post processed properly. In addition, we have also reported the Spearman correlation coefficients between accuracy and the total cost, which confirmed the advantage of MAUC over accuracy as expected.

In the experiments, we generated the cost matrices randomly for 20 different runs, and evaluated the performance of classifiers based on these cost matrices. This follows the \emph{de facto} standard procedure in the cost-sensitive learning field \cite{Domingos_1999}. However, to evaluate the average performance of base classifiers or the cost-sensitive learning methods, one needs to sample the cost matrices in $\mathbb{R}^{c\times{c}}$ space. For even a moderate size of $c$, this sampling space is too huge to sample sufficiently. Given no prior knowledge of the cost matrices, it seems very hard if not impossible to estimate the performance stably. We will explore the problem in the future. Although the results reported in this study suggest that simple calibration methods outperform existing sophisticated re-optimization methods, it is possible that other re-optimization methods or another formulation of the re-optimization problem could perform better, e.g., considering new progresses in evolutionary computation field. This could be another future work.

\bibliographystyle{IEEEtran}
\bibliography{MyCollection}
\end{document}